\documentclass{article}
\usepackage{icassp2023_spconf,amsmath,graphicx}

\usepackage[mathscr]{eucal}

\usepackage{multirow}
\usepackage{makecell}
\usepackage{amssymb}
\usepackage{pifont}
\usepackage{nccmath}
\newcommand{\cmark}{\ding{51}}%
\newcommand{\xmark}{\ding{55}}%

\def\etal{{\em et al.~}}

\makeatother
\usepackage[normalem]{ulem}
\usepackage{xcolor}

\usepackage{hyperref} 

\usepackage{url}

\usepackage[backend=bibtex,citestyle=numeric-comp,bibstyle=ieee,sorting=none,defernumbers=true,giveninits=true,doi=false,isbn=false,url=false,eprint=false,minbibnames=1,maxbibnames=1]{biblatex}
\usepackage{tabularx}
\usepackage{booktabs}
\usepackage{multirow}
\usepackage{ragged2e}
\usepackage{bm}

\usepackage{hyphenat}
\usepackage{xurl}

\usepackage{boldline}

\newcolumntype{C}{>{\Centering\arraybackslash}X}

\newcolumntype{u}{>{\raggedright\hsize=.7\hsize}X}
\newcolumntype{t}{>{\Centering\hsize=.6\hsize}X}
\newcolumntype{s}{>{\Centering\hsize=.5\hsize}X}
\newcolumntype{o}{>{\Centering\hsize=.4\hsize}X}
\newcolumntype{k}{>{\Centering\hsize=.3\hsize}X}
\newcolumntype{y}{>{\Centering\hsize=.2\hsize}X}
\newcolumntype{z}{>{\Centering\hsize=.1\hsize}X}
\newcolumntype{v}{>{\raggedright\hsize=.6\hsize}X}
\newcolumntype{e}{>{\raggedright\hsize=.5\hsize}X}
\newcolumntype{j}{>{\raggedright\hsize=.35\hsize}X}
\newcolumntype{f}{>{\raggedright\hsize=.3\hsize}X}
\newcolumntype{h}{>{\raggedright\hsize=.2\hsize}X}
\newcolumntype{q}{>{\raggedright\hsize=.8\hsize}X}
\title{Auto-AVSR: Audio-Visual Speech Recognition with Automatic Labels}

\makeatletter

\def\@name{\emph{Pingchuan Ma$^{1}$,  Alexandros Haliassos$^{1}$, Adriana Fernandez-Lopez$^2$, Honglie Chen$^2$} \\ \emph{Stavros Petridis$^{1,2}$, Maja Pantic$^{1,2}$
    \thanks{Only non-Meta co-authors downloaded, accessed, and used the datasets. Only non-Meta authors conducted any of the dataset pre-processing (no dataset pre-processing took place on Meta’s servers or facilities). Code and trained models are available at: \url{https://github.com/mpc001/auto_avsr}
    }
    }
}

\makeatother

\address{$^1$Imperial College London, UK\\
$^2$Meta AI, UK}

\begin{document}

\ninept

\maketitle

\begin{abstract}
Audio-visual speech recognition has received a lot of attention due to its robustness against acoustic noise. Recently, the performance of automatic, visual, and audio-visual speech recognition (ASR, VSR, and AV-ASR, respectively) has been substantially improved, mainly due to the use of larger models and training sets. However, accurate labelling of datasets is time-consuming and expensive. Hence, in this work, we investigate the use of automatically-generated transcriptions of unlabelled datasets to increase the training set size. For this purpose, we use publicly-available pre-trained ASR models to automatically transcribe unlabelled datasets such as AVSpeech and VoxCeleb2. Then, we train ASR, VSR and AV-ASR models on the augmented training set, which consists of the LRS2 and LRS3 datasets as well as the additional automatically-transcribed data. We demonstrate that increasing the size of the training set, a recent trend in the literature, leads to reduced WER despite using noisy transcriptions. The proposed model achieves new state-of-the-art performance on AV-ASR on LRS2 and LRS3. In particular, it achieves a WER of~0.9\,\% on LRS3, a relative improvement of 30\,\% over the current state-of-the-art approach, and outperforms  methods that have been trained on non-publicly available datasets with~26 times more training data.
\end{abstract}
 
\section{Introduction}
In the human perceptual system, the visual and audio streams often complement each other, yielding a unified robust response. It is also known that using visual signals along with audio signals leads to higher model robustness than using a single modality, especially in the presence of high levels of acoustic noise~\cite{afouras2018deep, petridis2018audio, makino2019recurrent, DBLP:conf/interspeech/ShiHM22}.

In this paper, we tackle Audio-Visual Automatic Speech Recognition~(AV-ASR, or~AVSR), which aims to transcribe continuous spoken sentences from both audio and visual streams. Recent audio-~(ASR), video-~(VSR) and audio-visual-based models have relied heavily on large-scale and well-labelled transcriptions to achieve convincing performance. However, accurate transcriptions require manual labelling, which is time-consuming and prohibitively expensive. In order to address this issue, several works have been proposed to build advanced ASR and VSR models by leveraging large-scale unlabelled audio-visual datasets. Popular approaches pre-train ASR and VSR models using self-supervised learning, where the goal is to learn audio and visual representations from large unlabelled datasets~\cite{baevski2020wav2vec, ma21c_interspeech, DBLP:journals/corr/abs-2201-02184, hsu2021hubert}. The pre-trained models are then fine-tuned on smaller labelled datasets. Alternatively, another line of work solves the task using knowledge distillation~\cite{afouras2020asr, DBLP:conf/cvpr/RenDLHH21}. For example, Afouras \etal~\cite{afouras2020asr} use a pre-trained ASR model~-~acting as a teacher~-~to provide an extra supervisory signal to the target VSR model, where the goal is to force the posterior distribution of the VSR network to match the teacher. Similarly, Ren \etal~\cite{DBLP:conf/cvpr/RenDLHH21} further improve the performance of visual-only models by distilling knowledge from pre-trained models with multiple modalities.

In this work, we propose a different approach to leverage large unlabelled datasets which does not require a two-step training approach used in self-supervised learning (SSL) (but it can easily be combined with any state-of-the-art SSL approach).  In particular, we take advantage of the availability of good publicly-available pre-trained ASR models~\cite{radford2018improving, baevski2020wav2vec, hsu2021hubert, DBLP:journals/corr/abs-1909-09577} to automatically annotate large-scale audio-visual datasets. This approach is broadly related to self-training~\cite{DBLP:conf/cvpr/XieLHL20, DBLP:conf/icassp/Kahn0H20, DBLP:conf/interspeech/ParkZJHCLWL20}, where a model is first trained on annotated data and then used to generate pseudo-labels for the unlabelled data. A new model is trained with all the annotated data, and this process is repeated for a few iterations. This iterative process might be necessary in other domains for which a high-quality pre-trained model does not exist, but this is not the case for ASR, where accurate pre-trained models are relatively abundant. Thus, we sidestep the need for a costly iterative procedure. Moreover, we incorporate the automatically-transcribed unlabelled data into the training set rather than using the pre-trained ASR model for distillation~\cite{afouras2020asr}. As a result, we can easily train a large-scale AV-ASR system by simplifying the implementation and reducing both the computational and memory costs of using a teacher during training. We also find similarities with~\cite{makino2019recurrent, serdyuk2022transformer}, but instead of using owner-uploaded transcriptions or a production-quality ASR system, all models and datasets we use are publicly accessible. 

Our main contributions can be summarised as follows:~1) we automatically generate transcriptions for more than~2\,000 hours of videos by utilising publicly-available ASR models. We then train ASR, VSR and AV-ASR models with these transcriptions and achieve state-of-the-art performance on the LRS2 and LRS3 datasets. Concretely, the proposed approach leads to a WER of~0.9\% for AV-ASR on the LRS3 dataset, which outperforms models trained on much larger training sets;~2) We show that the accuracy of the pre-trained ASR models used to automatically transcribe the unlabelled datasets is not highly correlated with the performance of the ASR and VSR models trained with these transcriptions; 3) We observe that an increase in the number of hours of automatically-transcribed data used in the training set results in reduced WER, especially for the VSR models. On the other hand, the performance of the ASR models seems to saturate beyond~1500 hours.

\begin{figure*}[!t]
    \centering
    \includegraphics[width=1.8\columnwidth]{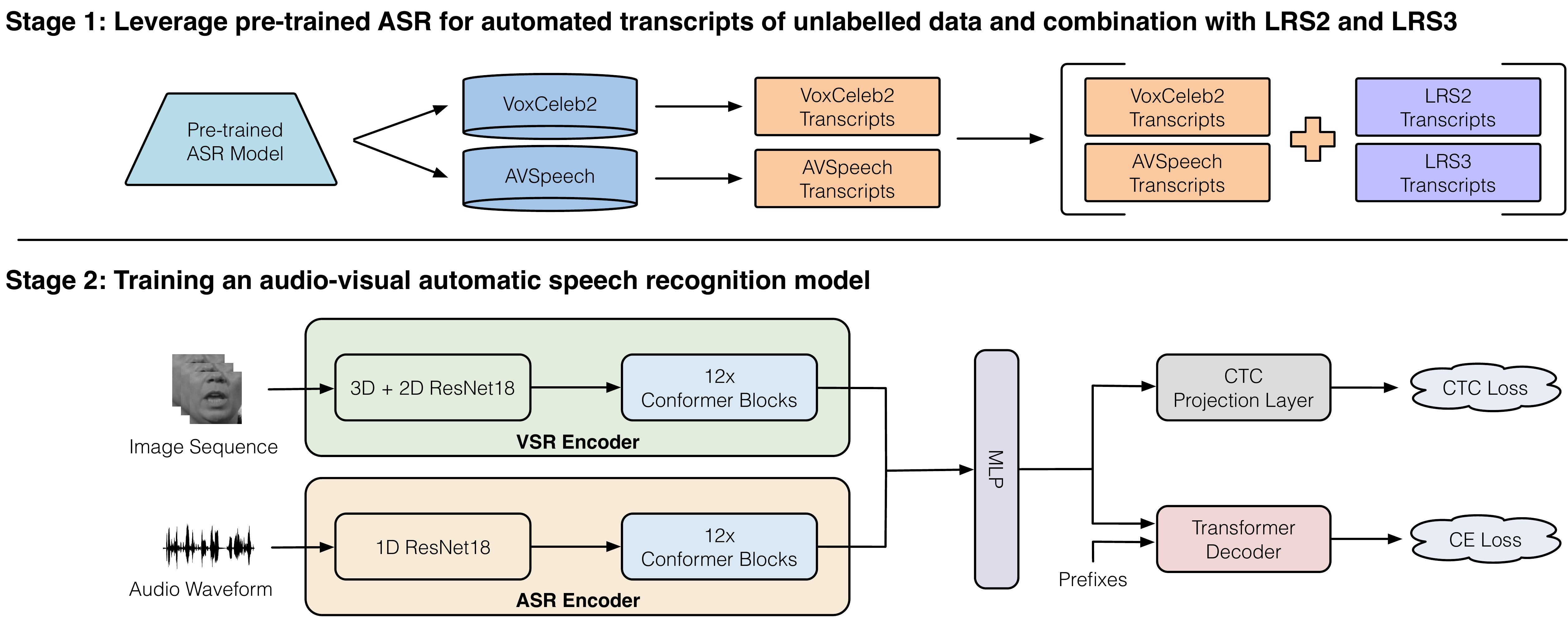}
    \caption{AV-ASR architecture overview.
    In the first stage, a pre-trained ASR model is leveraged to produce automatically-generated transcriptions for unlabelled audio-visual datasets. And then these unlabelled datasets are combined with the labelled training sets, including LRS2 and LRS3, for training. The frame rate of audio and visual features from the ASR and VSR encoders is 25 frames per second (fps).}
    \label{fig:av_architecture}
    \vspace{-4mm}
\end{figure*}
\vspace{-1mm}
\section{Auto-AVSR}
\subsection{Leveraging pre-trained models to produce automatically-generated transcriptions for unlabelled audio-visual datasets}
\label{sec:leveraging}
In order to investigate the impact of the size of training data on the performance of audio-only, visual-only and audio-visual models, we scale the training sources by including publicly-available audio-visual clips into the training set.
An overview of our label generation pipeline can be found at the top of Fig.~\ref{fig:av_architecture}. To be specific, audio waveforms from the unlabelled audio-visual datasets are fed into a pre-trained ASR model to produce automatic transcriptions.
For the purpose of this study, we use two unlabelled datasets: VoxCeleb2~\cite{DBLP:conf/interspeech/ChungNZ18} and AVSpeech~\cite{DBLP:journals/tog/EphratMLDWHFR18}. In particular, AVSpeech~\cite{DBLP:journals/tog/EphratMLDWHFR18} contains 4\,700 hours of YouTube video segments in multiple languages, and VoxCeleb2~\cite{DBLP:conf/interspeech/ChungNZ18} consists of 2\,300 hours of video segments from more than 6\,000 people. However, we are interested in training models in English. Thus, we use the VoxLingua107 language classifier~\cite{valk2021slt}
to filter the AVSpeech dataset, resulting in a total of 1\,323 hours; the list of English data we use for VoxCeleb2 is obtained from~\cite{DBLP:journals/corr/abs-2201-02184}, and comprises 1\,307 data hours. Next, we leverage publicly-available ASR models to produce automatically generated transcriptions. It is worth pointing out that our work facilitates reproduction and comparison since all datasets and models used are publicly accessible.

\vspace{-2mm}
\subsection{Automatic speech recognition models}
\label{sec:asr}
We investigate the impact of the automatic transcriptions given by four different ASR models on the performance of audio-only and visual-only models, i.e. Whisper~\cite{radford2018improving}, wav2vec2.0~\cite{baevski2020wav2vec}, Hidden unit BERT (HuBERT)~\cite{hsu2021hubert} and Conformer-Transducer~\cite{DBLP:journals/corr/abs-1909-09577, gulati2020conformer}. In particular, wav2vec\,2.0~\cite{baevski2020wav2vec} and HuBERT~\cite{hsu2021hubert} are self-supervised learning methods for learning speech representations from unlabelled audio. In contrast, Conformer-Transducer~\cite{DBLP:journals/corr/abs-1909-09577} is a conformer-based model trained with recurrent neural network transducers (RNN-T) loss that uses the NeMo ASRSET dataset, consisting of 12\,000 hours of English speech. Finally, Whisper~\cite{radford2018improving} is a transformer-based model~\cite{vaswani2017attention} trained with a total of 680\,000\,hours of labelled audio. In this work, we access the ASR models from the Hugging Face community, ``{nvidia/stt\_en\_conformer\_transducer\_xlarge}'', ``{facebook/hubert-large-ls960-ft}'',
``{facebook/wav2vec2-base-960h}'', and ``{openai/whisper-medium.en}'', respectively.

\vspace{-2mm}
\subsection{Architecture}
\label{sec:architecture}
We adopt the off-the-shelf architecture presented in \cite{DBLP:journals/corr/abs-2102-06657}, which has achieved state-of-the-art performance on the LRS2 and LRS3 datasets without the use of external data. The architecture is shown at the bottom of Fig.~\ref{fig:av_architecture}. In particular, the VSR front-end is based on a modified ResNet-18~\cite{he2016deep, stafylakis2017combining}, where the first layer is a spatio-temporal convolutional layer with a kernel size of 5$\times$7$\times$7 and a stride of 1$\times$2$\times$2. The temporal back-end, which follows the front-end, is a Conformer~\cite{gulati2020conformer}. Similarly, the ASR encoder consists of a~1D ResNet-18~\cite{DBLP:conf/icassp/PetridisSMCTP18} followed by a Conformer. The ASR and VSR encoder outputs are fused via a multi-layer perceptron (MLP). The rest of the network consists of a projection layer and a Transformer decoder for joint CTC/attention training~\cite{watanabe2017hybrid}.

\vspace{-1mm}
\section{Experimental Setup}
\subsection{Datasets}
We conduct experiments on LRS2~\cite{chung2017lip} and LRS3~\cite{afouras2018lrs3}, which are the two largest publicly available datasets for audio-visual speech recognition in English. LRS2, collected from BBC programs, contains~144\,482 video clips with a total of~225 hours. Specifically, the pre-training, training, validation and test set contains~96\,318~(195 hours),~45\,839~(28 hours),~1\,082~(0.6 hours) and~1\,243~(0.5 hours) video clips, respectively. LRS3 consists of~151\,819 video clips from TED talks with a total of 439 hours. It contains~118\,516~(408 hours),~31\,982~(30 hours) and~1\,321 clips~(0.9 hours) in the pre-training, training-validation, and test set, respectively. For training, we also use the English-speaking videos from AVSpeech (1\,323 hours) and VoxCeleb2 (1\,307 hours) as the additional training data together with automatically-generated transcriptions.

\vspace{-2mm}
\subsection{Pre-processing}
For the visual stream, we follow previous work~\cite{DBLP:journals/corr/abs-2102-06657} to pre-process the datasets. We crop the mouth region of interests (ROIs) using a bounding box of 96 $\times$ 96. Each frame is normalised by subtracting the mean and dividing by the standard deviation of the training set. For audio streams, we only perform $z$-normalisation per utterance.

\subsection{Implementation details}
For our audio- and visual-only ASR models, we use a ResNet-based front-end module pre-trained on LRW~\cite{ma2020towards}, followed by a Conformer encoder with 12 layers, 768 input dimensions, 3\,072 feed-forward dimensions, and 16 attention heads. The decoder is a 6-layer Transformer with the same dimensions and number of heads as the encoder, resulting in a total of 243.1\,M and 250.4\,M parameters for the audio- and visual-only models, respectively. More specifically, the ASR front-end, VSR front-end, Conformer back-end, Transformer decoder and the projection layer of the CTC have~3.9\,M,~11.2\,M,~170.9\,M,~64.5\,M and~3.9\,M parameters, respectively. For the audio-visual models, we concatenate the audio-visual encoder outputs and feed them to a~2-layer multi-layer perceptron~(MLP) with hidden and output sizes of~8\,192 and~768, respectively. 

For data augmentation, we apply horizontal flipping, random cropping, and adaptive time masking~\cite{ma2022visual} to the visual inputs, while we only use adaptive time masking for the audio stream. For both streams, we choose a number of masks that is proportional to the utterance length and a maximum masking length of up to 0.4 seconds. For the target vocabulary, we use SentencePiece~\cite{DBLP:conf/acl/Kudo18} subword units with a vocabulary size of 5\,000. We train the model for 75 epochs with the AdamW~\cite{loshchilov2019decoupled} optimizer, a cosine learning rate scheduler, and a warm-up of 5 epochs. The peak learning rate is 1e-3. The maximal number of frames in each batch is 1\,800 frames. Following~\cite{ma2022visual}, our visual-only models are incorporated with a transformer-based language model trained on a corpus of 166 million characters~$^\dagger$\let\thefootnote\relax\footnote{$^\dagger$ The corpus consits of the training sets of LibriSpeech ($960$\,h)~\cite{panayotov2015librispeech}, pre-training and training sets of LRS2~\cite{chung2017lip} and LRS3~\cite{afouras2018lrs3}, TED-LIUM 3~\cite{DBLP:conf/specom/HernandezNGTE18}, Voxforge (English) and Common Voice (English)~\cite{DBLP:conf/lrec/ArdilaBDKMHMSTW20}} whereas language models are not included for ASR and AV-ASR models since no further improvements are observed.

\vspace{-1mm}
\section{Results}

\begin{table}[!t]
\small
\centering
\renewcommand\arraystretch{0.8}
\begin{tabularx}{.99\linewidth}{v  y  y | y y}
\toprule
    \multirow{2}{*}[-0.1em]{\textbf{Method}} & \multicolumn{4}{c}{\textbf{WER [\%]}} \\
\cmidrule(lr){2-5}
 & A$^\dagger$ & A$^{\dagger\dagger}$ & V  & A \\
\midrule\midrule
CM-Transducer~\cite{DBLP:journals/corr/abs-1909-09577} & 1.62 &3.31 & 19.1 &0.99 \\
\midrule
HuBERT~\cite{hsu2021hubert} & 1.90 &6.87 & 19.8 &1.12 \\
\midrule
Wav2vec\,2.0~\cite{baevski2020wav2vec} & 3.40 &11.22 & 19.1 &1.06\\
\midrule
Whisper~\cite{radford2018improving} & 4.10 &1.81 & 19.0 &1.04 \\
\bottomrule
\end{tabularx}
\caption{
Impact of the pre-trained ASR models used to generate automatic transcriptions from unlabelled data on the performance of VSR/ASR models on the LRS3 dataset. $^\dagger$ and $^{\dagger\dagger}$ denote the word error rate (WER) reported on Librispeech test-clean set~\cite{panayotov2015librispeech} and LRS3 test set~\cite{afouras2018lrs3}, respectively. ``CM'' denotes Conformer. ``V'' and ``A'' denote the visual-only and audio-only models trained on LRW, LRS2, LRS3, VoxCeleb2 and AVSpeech (using the automatically-generated transcriptions from the corresponding pre-trained ASR model), with a total of~3\,448 hours.}
\label{tab:different_asr_models_lrs3}
\vspace{-5mm}
\end{table}
\vspace{-1mm}
\subsection{Do better Librispeech ASR models provide better transcriptions for VSR?}
Given that several publicly-available ASR models are available, we use performance on Librispeech as a criterion for model selection. We use models that have achieved state-of-the-art performance on the test-clean set of Librispeech, i.e., Conformer-Transducer~\cite{DBLP:journals/corr/abs-1909-09577} and HuBERT~\cite{hsu2021hubert}. We also use  ASR models that are widely used in the speech community, wav2vec\,2.0~\cite{baevski2020wav2vec} and Whisper~\cite{radford2018improving}.
Performance of ASR models on Librispeech clean-test set is shown in the first column of Table~\ref{tab:different_asr_models_lrs3}. 
Results of the ASR and VSR models trained with the automatically-generated transcriptions on the LRS3 dataset are shown in the third and fourth columns, respectively, of Table~\ref{tab:different_asr_models_lrs3}.
We observe that overall the WER on Librispeech is not highly correlated with the performance of the ASR and VSR models trained with the automatically-generated transcriptions from the corresponding pre-trained ASR models. The same conclusion is also true when we measure the WER on the LRS3 test. We show that using the transcriptions from most ASR models (i.e., wav2vec\,2.0~\cite{baevski2020wav2vec}, Whisper~\cite{radford2018improving}, and Conformer-Transducer~\cite{DBLP:journals/corr/abs-1909-09577}) results in very similar WER for both audio-only and visual-only models. The only exception is the use of automatically-generated transcriptions from HuBERT~\cite{hsu2021hubert} which results in slightly worse performance despite being one of the best performing models on Librispeech. 
In this work, we rely on the automatically-generated transcriptions from the Conformer-Transducer~\cite{DBLP:journals/corr/abs-1909-09577} since on average it leads to the best performance for both ASR and VSR models.

\vspace{-1mm}
\subsection{Impact of the number of hours of unlabelled data}
\begin{table}
\small
\centering
\renewcommand\arraystretch{0.8}
\begin{tabularx}{.99\linewidth}{t | c c c c c c }
\toprule
\textbf{P}& 0\,\% & 20\,\% & 40\,\% & 60\,\% & 80\,\% & 100\,\%
    \\
\textbf{U}& $\mathsf{0}$\,h &$\mathsf{526}$\,h &$\mathsf{1\,052}$\,h &$\mathsf{1\,578}$\,h &$\mathsf{2\,104}$\,h &$\mathsf{2\,630}$\,h
    \\
\textbf{T}& $\mathsf{818}$\,h &$\mathsf{1344}$\,h &$\mathsf{1\,870}$\,h &$\mathsf{2\,396}$\,h &$\mathsf{2\,922}$\,h &$\mathsf{3\,448}$\,h
    \\
\midrule\midrule
A & 1.5 & 1.3 & 1.3 & 1.1 & 1.0 &1.0 \\
\midrule
V & 33.0 & 26.6 & 23.6 & 21.9 & 20.0 &19.1 \\
\bottomrule
\end{tabularx}
\caption{
Impact of the size of additional training data (from AVSpeech and VoxCeleb2) on the WER (\%) of audio-only and visual-only models evaluated on LRS3. All models are initialised from a model pre-trained on LRW and trained on LRS2, LRS3 plus X\,\% hours of VoxCeleb2 and AVSpeech. ``\textbf{P}'' and ``\textbf{U}'' denote the amount of additional data in percentages and in hours, respectively. ``\textbf{T}'' denotes the total amount of training data (hours).}
\label{tab:percentage_experiments_on_lrs3}
\vspace{-5mm}
\end{table}
Table~\ref{tab:percentage_experiments_on_lrs3} shows the impact of varying the numbers of hours of unlabelled data on the performance of ASR and VSR models on LRS3. An absolute improvement of~1.7\,\% in WER is observed for VSR by using only labelled data from LRS2 and LRS3~(818 hours) compared to~\cite{ma2022visual}. This gain is likely due to the increase in model capacity. When including~20\,\%~(526 hours) of AVSpeech~\cite{DBLP:journals/tog/EphratMLDWHFR18} and VoxCeleb2~\cite{DBLP:conf/interspeech/ChungNZ18}, the performance of audio- and visual-only models can be further improved to 1.3\,\% and 26.6\,\% WER, respectively. Increasing further the number of training hours leads to a further reduction of the WER especially for the VSR model. This is in line with the recent trend observed in the literature~\cite{ma2022visual}, where using larger training sets substantially improves performance. In this experiment, we also show that the WER can be improved even by adding data that have been automatically transcribed and inevitably have noisy labels. We also notice that the improvement for the ASR model is marginal when using more than 1\,578 hours of unlabelled training data, indicating that the ASR performance may have saturated.

\begin{table}[!t]
\centering
\small
\renewcommand\arraystretch{0.9}
\begin{tabularx}{1.05\linewidth}{l y y y y}
\toprule
\textbf{Method} &\textbf{Type} &\textbf{Extra Data} &\textbf{Total Hours$^\dagger$} & \textbf{WER} (\%) \\ \midrule\midrule
MV-WAS~\cite{chung2017lip} &\multirow{13}{*}[-0.4em]{V} &\multirow{4}{*}[-0.1em]{\xmark} &\multirow{4}{*}[-0.1em]{223} &70.4 \\
TDNN~\cite{yu2020audio} & & & & 48.9 \\
CM-seq2seq~\cite{DBLP:journals/corr/abs-2102-06657} & & & &39.1 \\
CM-aux~\cite{ma2022visual} & & &  &32.9 \\
\cmidrule(lr){1-1} \cmidrule(lr){3-5}
CTC/Attention \cite{petridis2018audio} & &\multirow{9}{*}[-0.4em]{\cmark}  &380 &63.5 \\
KD\,+\,CTC  \cite{afouras2020asr} & & &995 &51.3 \\
KD-seq2seq~\cite{DBLP:conf/cvpr/RenDLHH21} & & &818 &49.2 \\
TM-seq2seq~\cite{afouras2018deep} & & &1\,391 &48.3 \\
CTC/Attention~\cite{pan2022leveraging} & & &60\,000 &43.2 \\
CM-aux~\cite{ma2022visual} & & &1\,459  &25.5 \\
VTP~\cite{prajwal2022sub} & & &2\,676 &22.6 \\
Ours& & &818 &\textbf{27.9} \\
Ours& & &3\,448 &\textbf{14.6} \\
\midrule
TDNN~\cite{yu2020audio}  &\multirow{5}{*}[-0.4em]{A} &\multirow{2}{*}[-0.1em]{\xmark} &\multirow{2}{*}[-0.1em]{223} &6.7 \\
CM-seq2seq~\cite{DBLP:journals/corr/abs-2102-06657} & & & &4.3 \\
\cmidrule(lr){1-1} \cmidrule(lr){3-5}
CTC/Attention~\cite{pan2022leveraging} & &\multirow{3}{*}[-0.1em]{\cmark} &60\,000 &2.7 \\
Ours& & &818 &\textbf{2.6} \\
Ours& & &3\,448 &\textbf{1.5} \\
\midrule
TDNN~\cite{yu2020audio}  &\multirow{6}{*}[-0.1em]{A+V} &\multirow{2}{*}[-0.1em]{\xmark} &\multirow{2}{*}[-0.1em]{223} &5.9 \\
CM-seq2seq~\cite{DBLP:journals/corr/abs-2102-06657} & & & &4.2 \\
\cmidrule(lr){1-1} \cmidrule(lr){3-5}
TM-seq2seq~\cite{afouras2018deep} & &\multirow{4}{*}[-0.1em]{\cmark} &1\,391 &8.3 \\
CTC/Attention \cite{petridis2018audio} & &  &380 &7.0 \\
CM-seq2seq~\cite{DBLP:journals/corr/abs-2102-06657}  & & &380 &3.9 \\
Ours& & &3\,448 &\textbf{1.5} \\
\bottomrule
\end{tabularx}
\caption{WER (\%) of our audio-only, visual-only and audio-visual models on the LRS2 dataset. $^\dagger$ The total hours are counted by including the datasets used for both pre-training and training. Our model trained on 818 hours uses LRW, LRS2 and LRS3. Our model trained on 3\,448 hours uses LRW, LRS2, LRS3, VoxCeleb2 and AVSpeech.}
\vspace{-4mm}
\label{tab:sota lrs2}
\end{table}
\begin{table}[!t]
\centering
\small
\renewcommand\arraystretch{0.9}
\begin{tabularx}{1.05\linewidth}{l y y y y}
\toprule
\textbf{Method} &\textbf{Type} &\textbf{Extra Data} &\textbf{Total Hours$^\ddagger$} & \textbf{WER} (\%) \\ \midrule\midrule
CM-seq2seq~\cite{DBLP:journals/corr/abs-2102-06657} &\multirow{13}{*}[-0.4em]{V} &\multirow{3}{*}[-0.2em]{\xmark} &\multirow{3}{*}[-0.1em]{438} &46.9 \\
CM-aux~\cite{ma2022visual} & & &  &37.9 \\
Ours & & &  &\textbf{36.3} \\
\cmidrule(lr){1-1}\cmidrule(lr){3-5}
KD\,+\,CTC~\cite{afouras2020asr} & &\multirow{10}{*}[-0.4em]{\cmark} &772 &59.8 \\
KD-seq2seq~\cite{DBLP:conf/cvpr/RenDLHH21} & & &818 &59.0 \\
TM-seq2seq~\cite{afouras2018deep} & & &1\,362 &58.9 \\
AVHuBERT~\cite{DBLP:journals/corr/abs-2201-02184} & & &1\,759 & 26.9 \\
RNN-T~\cite{makino2019recurrent} & & &31\,000 &33.6 \\
VTP~\cite{prajwal2022sub} & & &2\,676 &30.7 \\
ViT3D-CM~\cite{serdyuk2022transformer} & & &90\,000 &17.0 \\
Ours & & &818 &\textbf{33.0} \\
Ours& & &1\,902 &\textbf{23.5} \\
Ours & & &3\,448 &\textbf{19.1} \\
\midrule
CM-seq2seq~\cite{DBLP:journals/corr/abs-2102-06657} &\multirow{6}{*}[-0.4em]{A} &\multirow{1}{*}[-0.1em]{\xmark} &438 &2.3 \\
\cmidrule(lr){1-1}\cmidrule(lr){3-5}
RNN-T~\cite{makino2019recurrent} & &\multirow{5}{*}[-0.1em]{\cmark} &31\,000 &4.5 \\
AV-HuBERT~\cite{DBLP:journals/corr/abs-2201-02184}  & & &1\,759 &1.3 \\
Ours& & &818 &\textbf{1.5} \\
Ours& & &1\,902 &\textbf{1.0} \\
Ours& & &3\,448 &\textbf{1.0} \\
\midrule
CM-seq2seq~\cite{DBLP:journals/corr/abs-2102-06657} &\multirow{6}{*}[-0.4em]{A+V} &\multirow{1}{*}[-0.1em]{\xmark} &438 &2.3 \\
\cmidrule(lr){1-1}\cmidrule(lr){3-5}
RNN-T~\cite{makino2019recurrent} & &\multirow{5}{*}[-0.1em]{\cmark} &31\,000 &4.8 \\
AV-HuBERT~\cite{DBLP:journals/corr/abs-2201-02184} & & &1\,759 &1.4 \\
ViT3D-CM~\cite{serdyuk2022transformer} & & &90\,000 &1.6 \\
Ours& & &1\,902 &\textbf{1.0} \\
Ours& & &3\,448 &\textbf{0.9} \\
\bottomrule
\end{tabularx}
\caption{WER (\%) of our audio-only, visual-only and audio-visual models on the LRS3 dataset. $^\ddagger$ The total hours are counted by including the datasets used for both pre-training and training. Our model trained on 818 hours uses LRW, LRS2 and LRS3. Our model trained on 1\,902 hours uses LRW, LRS3 and VoxCeleb2. Our model trained on 3\,448 hours uses LRW, LRS2, LRS3, VoxCeleb2 and AVSpeech.}
\label{tab:sota lrs3}
\vspace{-1mm}
\end{table}
\vspace{-1mm}
\subsection{Comparison with the state-of-the-art}
Results on LRS2 and LRS3 are presented in Tables~\ref{tab:sota lrs2} and~\ref{tab:sota lrs3}, respectively. For LRS2, it is clear that our visual-only, audio-only and audio-visual models further push the state-of-the-art performance to a WER of~14.6\,\%, 1.5\,\% and 1.5\,\% respectively. For LRS3, the best visual-only model has a WER of 19.1\,\%, which is outperformed only by~\cite{serdyuk2022transformer} (17.0\,\% WER) which uses 26$\times$ more training data.
Similarly, our audio-only model establishes a new state-of-the-art~\cite{DBLP:journals/corr/abs-2201-02184} by achieving a WER of 1.0\,\% when using 1\,921 hours of training data from LRW, LRS3 and VoxCeleb2 datasets. However, when further introducing AVSpeech for training, no further improvement is observed, suggesting that the ASR performance may have reached saturation. State-of-the-art performance is also achieved for AV-ASR with a WER of 0.9\,\%.

\vspace{-1mm}
\begin{table}[!bt]
\centering
\small
\renewcommand\arraystretch{0.8}
\begin{tabularx}{.99\linewidth}{o o | c c c c c  }
\toprule
\multirow{2}{*}{\textbf{Type}}&\multirow{2}{*}{\textbf{Noise}} &\multicolumn{5}{c}{\textbf{SNR levels [dB]}}
\\
& & 12.5 & 7.5 & 2.5 & -2.5 & -7.5
\\
\midrule\midrule
A &\multirow{2}{*}{Babble$^{\ddagger}$} &1.1 & 1.2 & 1.6 & 2.7 & 8.3
\\
\cmidrule(lr){1-1}\cmidrule(lr){3-7}
A+V & &1.0 &1.0 &1.5 &2.2 &5.6
\\
\midrule
A &\multirow{2}{*}{Pink} &1.4 & 1.9 & 4.3 & 13.1 & 56.8
\\
\cmidrule(lr){1-1}\cmidrule(lr){3-7}
A+V & &1.2 &1.4 &2.3 &6.0 &16.2
\\
\midrule
A &\multirow{2}{*}{White} &2.1 &4.0 &10.4 &30.2 &88.9
\\
\cmidrule(lr){1-1}\cmidrule(lr){3-7}
A+V & &1.4 &2.3 &4.3 &9.5 &24.2
\\
\bottomrule
\end{tabularx}
\caption{WER (\%) of our audio-only and audio-visual models as a function of the noise levels on the LRS3 dataset. The babble noise from the NOISEX dataset~\cite{DBLP:journals/speech/VargaS93} is used for training while one of SNR levels from [-5\,dB, 0\,dB, 5\,dB, 10\,dB, 15\,dB, 20\,dB, $\infty$ dB] is selected with a uniform distribution. For testing, the pink and white noise from the Speech Commands dataset~\cite{DBLP:journals/corr/abs-1804-03209} is added to the raw audio waveforms with a specific SNR level. $\ddagger$ denotes the noise type used in both training and test set.}
\label{tab: noise experiments}
\vspace{-6 mm}
\end{table}

\subsection{Noise experiments}
Results of ASR and AV-ASR models, when tested with different acoustic noise levels, are shown in Table.~\ref{tab: noise experiments}.
During training we use the babble noise from the NOISEX dataset~\cite{DBLP:journals/speech/VargaS93}, while the SNR level is selected from [-5\,dB, 0\,dB, 5\,dB, 10\,dB, 15\,dB, 20\,dB, $\infty$ dB] with a uniform distribution. For evaluation, we test three types of noise:  babble noise~\cite{DBLP:journals/speech/VargaS93}, pink and white noise from the Speech Commands dataset~\cite{DBLP:journals/corr/abs-1804-03209}.
We show that, overall, the results are consistent with those presented in~\cite{afouras2018deep, DBLP:journals/corr/abs-2102-06657, makino2019recurrent, DBLP:conf/interspeech/ShiHM22}, i.e. the performance of audio-only models is closer to the audio-visual counterpart in the presence of low levels of noise, whereas the performance gap becomes larger as the noise levels increase. We notice that when using babble noise for evaluation, the performance of either audio-only or audio-visual models has a WER lower than 10\,\% at -7.5\,dB. This is likely mainly a consequence of the overlapping noise type in the training and testing phases (despite mismatched levels of noise).

\vspace{-1mm}
\section{Conclusions}
In this work, we propose a simple and efficient method for scaling up audio-visual data for speech recognition. We present a detailed study on the performance of LRS3 in terms of the amount of unlabelled training data. By leveraging publicly-available ASR models to produce automatically-generated transcriptions, we train an AV-ASR system and achieve state-of-the-art performance on both publicly-available audio-visual benchmarks, LRS2 and LRS3. Furthermore, we show that our audio-visual model is more robust against different levels of noise than its audio-only counterpart.

\newpage
\AtNextBibliography{\small}
\section{References}
\begingroup
\printbibliography[heading=none]

@string{icassp = "ICASSP"}

@string{interspeech = "Interspeech"}

@string{asru = "ASRU"}

@string{lrec = "LREC"}

@string{ieee-taslp = "IEEE Trans. Audio, Speech, Lang. Process."}

@string{pami = "IEEE TPAMI"}

@string{iclr = "ICLR"}

@string{cvpr = "CVPR"}

@string{acl = "ACL"}

@string{nips = "NIPS"}

@string{slt = "SLT"}

@string{specom = "SPECOM"}

@article{afouras2018deep,
  title={Deep audio-visual speech recognition},
  author={Afouras, T. and Chung, J. S. and Senior, A. and Vinyals, O. and Zisserman, A.},
  journal=pami,
  year={2018},
  page={1},
  doi={10.1109/TPAMI.2018.2889052}
}

@INPROCEEDINGS{petridis2018audio,
  author={S. {Petridis} and T. {Stafylakis} and P. {Ma} and G. {Tzimiropoulos} and M. {Pantic}},
  title={Audio-Visual Speech Recognition with a Hybrid {CTC}/Attention Architecture},
  booktitle=slt,
  year={2018},
  volume={},
  number={},
  pages={513-520},
  doi={10.1109/SLT.2018.8639643}}

@INPROCEEDINGS{makino2019recurrent,
  title={Recurrent neural network transducer for audio-visual speech recognition},
  author={Makino, T. and Liao, H. and Assael, Y. and Shillingford, B. and Garcia, B. and Braga, O. and Siohan, O.},
  booktitle=asru,
  year={2019},
  pages={905-912},
  doi={10.1109/ASRU46091.2019.9004036},
}

@inproceedings{baevski2020wav2vec,
  title={wav2vec 2.0: A framework for self-supervised learning of speech representations},
  author={Baevski, Alexei and Zhou, Yuhao and Mohamed, Abdelrahman and Auli, Michael},
  booktitle=nips,
  volume={33},
  pages={12449--12460},
  year={2020}
}

@inproceedings{ma21c_interspeech,
  author={Pingchuan Ma and Rodrigo Mira and Stavros Petridis and Björn W. Schuller and Maja Pantic},
  title={{Li{R}A: Learning visual speech representations from audio through self-supervision}},
  year=2021,
  booktitle=interspeech,
  pages={3011--3015},
  doi={10.21437/Interspeech.2021-1360},
  organization={},
}

@inproceedings{DBLP:journals/corr/abs-2201-02184,
  author    = {Bowen Shi and
               Wei{-}Ning Hsu and
               Kushal Lakhotia and
               Abdelrahman Mohamed},
  title     = {Learning Audio-Visual Speech Representation by Masked Multimodal Cluster
               Prediction},
  booktitle = iclr,
  publisher = {},
  year      = {2022},
  url       = {https://openreview.net/forum?id=Z1Qlm11uOM},
  bibsource = {dblp computer science bibliography, https://dblp.org}
}

@article{hsu2021hubert,
  title={Hubert: Self-supervised speech representation learning by masked prediction of hidden units},
  author={Hsu, Wei-Ning and Bolte, Benjamin and Tsai, Yao-Hung Hubert and Lakhotia, Kushal and Salakhutdinov, Ruslan and Mohamed, Abdelrahman},
  journal=ieee-taslp,
  volume={29},
  pages={3451--3460},
  year={2021},
  publisher={}
}

@INPROCEEDINGS{afouras2020asr,
  title={{ASR} is all you need: Cross-modal distillation for lip reading},
  author={Afouras, T. and Chung, J. S. and Zisserman, A.},
  booktitle=icassp,
  year={2020},
  pages={2143-2147},
  doi={10.1109/ICASSP40776.2020.9054253}
}

@inproceedings{DBLP:conf/cvpr/RenDLHH21,
  author    = {Sucheng Ren and
               Yong Du and
               Jianming Lv and
               Guoqiang Han and
               Shengfeng He},
  title     = {Learning From the Master: Distilling Cross-Modal Advanced Knowledge
               for Lip Reading},
  booktitle = cvpr,
  pages     = {13325--13333},
  publisher = {},
  year      = {2021},
  url       = {https://openaccess.thecvf.com/content/CVPR2021/html/Ren\_Learning\_From\_the\_Master\_Distilling\_Cross-Modal\_Advanced\_Knowledge\_for\_Lip\_CVPR\_2021\_paper.html},
  doi       = {10.1109/CVPR46437.2021.01312},
  bibsource = {dblp computer science bibliography, https://dblp.org}
}

@misc{radford2018improving,
  title={Introducing Whisper},
  author={Alec Radford and Jong Wook Kim and Christine McLeavey Payne and Pamela Mishkin and Tao Xu and Greg Brockman and Ilya Sutskever},
  year={2022},
  howpublished = "\url{https://openai.com/blog/whisper/}",
  note = "[Online; accessed 18-October-2022]"
}

@article{DBLP:journals/corr/abs-1909-09577,
  title={Nemo: a toolkit for building ai applications using neural modules},
  author={Kuchaiev, Oleksii and Li, Jason and Nguyen, Huyen and Hrinchuk, Oleksii and Leary, Ryan and Ginsburg, Boris and Kriman, Samuel and Beliaev, Stanislav and Lavrukhin, Vitaly and Cook, Jack and others},
  journal={arXiv preprint arXiv:1909.09577},
  year={2019}
}

@inproceedings{DBLP:conf/cvpr/XieLHL20,
  author    = {Qizhe Xie and
               Minh{-}Thang Luong and
               Eduard H. Hovy and
               Quoc V. Le},
  title     = {Self-Training With Noisy Student Improves ImageNet Classification},
  booktitle = cvpr,
  pages     = {10684--10695},
  publisher = {},
  year      = {2020},
  url       = {https://openaccess.thecvf.com/content\_CVPR\_2020/html/Xie\_Self-Training\_With\_Noisy\_Student\_Improves\_ImageNet\_Classification\_CVPR\_2020\_paper.html},
  doi       = {10.1109/CVPR42600.2020.01070},
  bibsource = {dblp computer science bibliography, https://dblp.org}
}

@inproceedings{DBLP:conf/icassp/Kahn0H20,
  author    = {Jacob Kahn and
               Ann Lee and
               Awni Y. Hannun},
  title     = {Self-Training for End-to-End Speech Recognition},
  booktitle = icassp,
  pages     = {7084--7088},
  publisher = {},
  year      = {2020},
  url       = {https://doi.org/10.1109/ICASSP40776.2020.9054295},
  doi       = {10.1109/ICASSP40776.2020.9054295},
  bibsource = {dblp computer science bibliography, https://dblp.org}
}

@inproceedings{serdyuk2022transformer,
  title={Transformer-Based Video Front-Ends for Audio-Visual Speech Recognition for Single and Multi-Person Video},
  author={Serdyuk, Dmitriy and Braga, Otavio and Siohan, Olivier},
  booktitle=interspeech,
  pages={2833--2837},
  year={2022}
}

@InProceedings{DBLP:conf/interspeech/ChungNZ18,
  author    = {Joon Son Chung and
               Arsha Nagrani and
               Andrew Zisserman},
  title     = {VoxCeleb2: Deep Speaker Recognition},
  booktitle = interspeech,
  pages     = {1086--1090},
  year      = {2018},
}

@article{DBLP:journals/tog/EphratMLDWHFR18,
  author    = {Ariel Ephrat and
               Inbar Mosseri and
               Oran Lang and
               Tali Dekel and
               Kevin Wilson and
               Avinatan Hassidim and
               William T. Freeman and
               Michael Rubinstein},
  title     = {Looking to listen at the cocktail party: a speaker-independent audio-visual
               model for speech separation},
  journal   = {{ACM} Transactions on Graphics},
  volume    = {37},
  number    = {4},
  pages     = {112:1--112:11},
  year      = {2018},
  url       = {},
  doi       = {10.1145/3197517.3201357},
  bibsource = {dblp computer science bibliography, https://dblp.org}
}

@inproceedings{valk2021slt,
  title={{VoxLingua107}: a Dataset for Spoken Language Recognition},
  author={J{\"o}rgen Valk and Tanel Alum{\"a}e},
  booktitle=slt,
  year={2021},
}

@INPROCEEDINGS{vaswani2017attention,
  title={Attention is all you need},
  author={Vaswani, A. and Shazeer, N. and Parmar, N. and Uszkoreit, J. and Jones, L. and Gomez, A.~N and Kaiser, {\L}. and Polosukhin, I.},
  booktitle=nips,
  year={2017},
  pages={6000-6010}
}

@inproceedings{DBLP:journals/corr/abs-2102-06657,
    author={Ma, Pingchuan and Petridis, Stavros and Pantic, Maja},
    title={End-To-End Audio-Visual Speech Recognition with Conformers},
    booktitle=icassp, 
    year={2021},
    volume={},
    number={},
    pages={7613-7617},
    organization={},
    doi={10.1109/ICASSP39728.2021.9414567}
}

@INPROCEEDINGS{he2016deep,
  title={Deep residual learning for image recognition},
  author={He, K. and Zhang, X. and Ren, S. and Sun, J.},
  booktitle=cvpr,
  year={2016},
  pages={770-778}
}

@INPROCEEDINGS{stafylakis2017combining,
  title={Combining Residual Networks with {LSTM}s for Lipreading},
  author={Stafylakis, T. and Tzimiropoulos, G.},
  booktitle=interspeech,
  volume={9},
  pages={3652--3656},
  year={2017}
}

@INPROCEEDINGS{gulati2020conformer,
  title={Conformer: Convolution-augmented Transformer for Speech Recognition},
  author={A. Gulati and J. Qin and C. Chiu and N. Parmar and Y. Zhang and J. Yu and W. Han and S. Wang and Z. Zhang and Y. Wu and others},
  booktitle=interspeech,
  pages={5036--5040},
  year={2020}
}

@article{watanabe2017hybrid,
  title={Hybrid {CTC}/attention architecture for end-to-end speech recognition},
  author={S. Watanabe and T. Hori and S. Kim and J. R Hershey and T. Hayashi},
  journal={IEEE J. Sel. Top. Signal Process.},
  year={2017},
  volume={11},
  number={8},
  pages={1240-1253},
  doi={10.1109/JSTSP.2017.2763455}
}

@INPROCEEDINGS{chung2017lip,
  title={Lip reading sentences in the wild},
  author={Chung, J. S. and Senior, A. and Vinyals, O. and Zisserman, A.},
  booktitle=cvpr,
  year={2017},
  pages={3444-3453}
}

@article{afouras2018lrs3,
  title={{LRS3-TED}: a large-scale dataset for visual speech recognition},
  author={Afouras, Triantafyllos and Chung, Joon Son and Zisserman, Andrew},
  journal={arXiv preprint arXiv:1809.00496},
  year={2018}
}

@article{ma2022visual,
  title={{Visual Speech Recognition for Multiple Languages in the Wild}},
  author={Ma, Pingchuan and Petridis, Stavros and Pantic, Maja},
  journal={{Nature Machine Intelligence}},
  doi={https://doi.org/10.1038/s42256-022-00550-z},
  pages={930--939},
  year={2022}
}

@inproceedings{DBLP:conf/acl/Kudo18,
  author    = {Taku Kudo},
  title     = {Subword Regularization: Improving Neural Network Translation Models
               with Multiple Subword Candidates},
  booktitle = acl,
  pages     = {66--75},
  publisher = {},
  year      = {2018},
  url       = {https://aclanthology.org/P18-1007/},
  doi       = {10.18653/v1/P18-1007},
  bibsource = {dblp computer science bibliography, https://dblp.org}
}

@inproceedings{loshchilov2019decoupled,
  author    = {Ilya Loshchilov and
               Frank Hutter},
  title     = {Decoupled Weight Decay Regularization},
  booktitle = iclr,
  publisher = {},
  year      = {2019},
  url       = {https://openreview.net/forum?id=Bkg6RiCqY7},
  bibsource = {dblp computer science bibliography, https://dblp.org}
}

@INPROCEEDINGS{panayotov2015librispeech,
  author={V. {Panayotov} and G. {Chen} and D. {Povey} and S. {Khudanpur}},
  booktitle=icassp, 
  title={Librispeech: An {ASR} corpus based on public domain audio books},
  year={2015},
  volume={},
  number={},
  pages={5206-5210},
  doi={10.1109/ICASSP.2015.7178964}
}

@INPROCEEDINGS{yu2020audio,
  author={J. {Yu} and S. {Zhang} and J. {Wu} and S. {Ghorbani} and B. {Wu} and S. {Kang} and S. {Liu} and X. {Liu} and H. {Meng} and D. {Yu}},
  booktitle=icassp, 
  title={Audio-Visual Recognition of Overlapped Speech for the {LRS2} Dataset}, 
  year={2020},
  volume={},
  number={},
  pages={6984-6988},
  doi={10.1109/ICASSP40776.2020.9054127}
}

@inproceedings{pan2022leveraging,
  title={Leveraging Uni-Modal Self-Supervised Learning for Multimodal Audio-Visual Speech Recognition},
  author={Pan, Xichen and Chen, Peiyu and Gong, Yichen and Zhou, Helong and Wang, Xinbing and Lin, Zhouhan},
  booktitle=acl,
  year={2022},
  pages={4491–4503}
}

@inproceedings{prajwal2022sub,
  title={Sub-word level lip reading with visual attention},
  author={Prajwal, KR and Afouras, Triantafyllos and Zisserman, Andrew},
  booktitle=cvpr,
  pages={5162--5172},
  year={2022}
}

@article{DBLP:journals/speech/VargaS93,
  author    = {Andrew Varga and
               Herman J. M. Steeneken},
  title     = {Assessment for automatic speech recognition: {II.} {NOISEX-92:} {A}
               database and an experiment to study the effect of additive noise on
               speech recognition systems},
  journal   = {Speech Commun.},
  volume    = {12},
  number    = {3},
  pages     = {247--251},
  year      = {1993},
  url       = {https://doi.org/10.1016/0167-6393(93)90095-3},
  doi       = {10.1016/0167-6393(93)90095-3},
  bibsource = {dblp computer science bibliography, https://dblp.org}
}

@article{DBLP:journals/corr/abs-1804-03209,
  author    = {Pete Warden},
  title     = {Speech Commands: {A} Dataset for Limited-Vocabulary Speech Recognition},
  journal   = {CoRR},
  volume    = {abs/1804.03209},
  year      = {2018},
  url       = {http://arxiv.org/abs/1804.03209},
  eprinttype = {arXiv},
  eprint    = {1804.03209},
  bibsource = {dblp computer science bibliography, https://dblp.org}
}

@inproceedings{DBLP:conf/interspeech/ParkZJHCLWL20,
  author    = {Daniel S. Park and
               Yu Zhang and
               Ye Jia and
               Wei Han and
               Chung{-}Cheng Chiu and
               Bo Li and
               Yonghui Wu and
               Quoc V. Le},
  title     = {Improved Noisy Student Training for Automatic Speech Recognition},
  booktitle = interspeech,
  pages     = {2817--2821},
  publisher = {},
  year      = {2020},
  url       = {https://doi.org/10.21437/Interspeech.2020-1470},
  doi       = {10.21437/Interspeech.2020-1470},
  bibsource = {dblp computer science bibliography, https://dblp.org}
}

@inproceedings{DBLP:conf/icassp/PetridisSMCTP18,
  author    = {Stavros Petridis and
               Themos Stafylakis and
               Pingchuan Ma and
               Feipeng Cai and
               Georgios Tzimiropoulos and
               Maja Pantic},
  title     = {End-to-End Audiovisual Speech Recognition},
  booktitle = icassp,
  pages     = {6548--6552},
  publisher = {},
  year      = {2018},
  url       = {https://doi.org/10.1109/ICASSP.2018.8461326},
  doi       = {10.1109/ICASSP.2018.8461326},
  bibsource = {dblp computer science bibliography, https://dblp.org}
}

@inproceedings{DBLP:conf/interspeech/ShiHM22,
  author    = {Bowen Shi and
               Wei{-}Ning Hsu and
               Abdelrahman Mohamed},
  title     = {Robust Self-Supervised Audio-Visual Speech Recognition},
  booktitle = interspeech,
  pages     = {2118--2122},
  publisher = {},
  year      = {2022},
  url       = {https://doi.org/10.21437/Interspeech.2022-99},
  doi       = {10.21437/Interspeech.2022-99},
  bibsource = {dblp computer science bibliography, https://dblp.org}
}

@inproceedings{DBLP:conf/specom/HernandezNGTE18,
  author    = {Fran{\c{c}}ois Hernandez and
               Vincent Nguyen and
               Sahar Ghannay and
               Natalia A. Tomashenko and
               Yannick Est{\`{e}}ve},
  title     = {{TED-LIUM} 3: Twice as Much Data and Corpus Repartition for Experiments
               on Speaker Adaptation},
  booktitle = specom,
  volume    = {11096},
  pages     = {198--208},
  year      = {2018},
  url       = {},
  doi       = {10.1007/978-3-319-99579-3\_21},
  bibsource = {dblp computer science bibliography, https://dblp.org}
}

@inproceedings{dblp:conf/lrec/ardilabdkmhmstw20,
    title = "Common Voice: A Massively-Multilingual Speech Corpus",
    author = "Ardila, Rosana  and
      Branson, Megan  and
      Davis, Kelly  and
      Kohler, Michael  and
      Meyer, Josh  and
      Henretty, Michael  and
      Morais, Reuben  and
      Saunders, Lindsay  and
      Tyers, Francis  and
      Weber, Gregor",
    booktitle = lrec,
    year = "2020",
    url = "",
    pages = "4218--4222",
    language = "English",
    ISBN = "",
}

@INPROCEEDINGS{ma2020towards,
  author={Ma, Pingchuan and Martinez, Brais and Petridis, Stavros and Pantic, Maja},
  booktitle=icassp,
  title={Towards Practical Lipreading with Distilled and Efficient Models},
  year={2021},
  pages={7608-7612},
  doi={10.1109/ICASSP39728.2021.9415063}
}
\endgroup

\end{document}